# The Danger Theory and Its Application to Artificial Immune Systems

Uwe Aickelin

School of Computer Science

University of Nottingham

NG8 1BB   UK

uxa@cs.nott.ac.uk

Steve Cayzer

Hewlett-Packard Laboratories

Filton Road

Bristol

BS12 6QZ

Steve_Cayzer@hp.com

## Abstract

Over the last decade, a new idea challenging the classical self-non-self viewpoint has become popular amongst immunologists. It is called the Danger Theory. In this conceptual paper, we look at this theory from the perspective of Artificial Immune System practitioners. An overview of the Danger Theory is presented with particular emphasis on analogies in the Artificial Immune Systems world. A number of potential application areas are then used to provide a framing for a critical assessment of the concept, and its relevance for Artificial Immune Systems.

## 1   INTRODUCTION

Over the last decade, a new theory has become popular amongst immunologists. It is called the Danger Theory, and its chief advocate is Matzinger [18], [19] and [20]. A number of advantages are claimed for this theory; not least that it provides a method of 'grounding' the immune response. The theory is not complete, and there are some doubts about how much it actually changes behaviour and / or structure. Nevertheless, the theory contains enough potentially interesting ideas to make it worth assessing its relevance to Artificial Immune Systems.

It should be noted that we do not intend to defend this theory, which is still controversial [21]. Rather we are interested in its merits for Artificial Immune System applications and hence its actual existence in the humoral immune system is of little importance to us. Our question is: Can it help us build better Artificial Immune Systems?

Few other Artificial Immune System practitioners are aware of the Danger Theory, notable exceptions being Burgess [5] and Willamson [22]. Hence, this is the first paper that deals directly with the Danger Theory, and it is the authors' intention that this paper stimulates discussion in our research community.

In the next section, we provide an overview of the Danger Theory, pointing out, where appropriate, some analogies in current Artificial Immune System models. We then assess the relevance of the theory for Artificial Immune System security applications, which is probably the most obvious application area for the danger model. Other Artificial Immune System application areas are also considered. Finally, we draw some preliminary conclusions about the potential of the Danger concept.

## 2   THE DANGER THEORY

The immune system is commonly thought to work at three levels: External barriers (skin, mucus), innate immunity and the acquired or adaptive immune system. As part of the third and most complex level, B-Lymphocytes secrete specific antibodies that recognise and react to stimuli. It is this pattern matching between antibodies and antigens that lies at the heart of most Artificial Immune System implementations. Another type of cell, the T (killer) lymphocyte, is also important in different types of immune reactions. Although not usually present in Artificial Immune System models, the behaviour of this cell is implicated in the Danger model and so it is included here. From the Artificial Immune System practitioner's point of view, the T killer cells match stimuli in much the same way as antibodies do.

However, it is not simply a question of matching in the humoral immune system. It is fundamental that only the 'correct' cells are matched as otherwise this could lead to a self-destructive autoimmune reaction. Classical immunology [12] stipulates that an immune response is triggered when the body encounters something non-self or foreign. It is not yet fully understood how this self-non-self discrimination is achieved, but many immunologists believe that the difference between them is learnt early in life. In particular it is thought that the maturation process plays an important role to achieve self-tolerance by eliminating those T and B cells that react to self. In addition, a 'confirmation' signal is required; that is, for either B cell or T (killer) cell activation, a T (helper) lymphocyte must also be activated. This dual activation is further protection against the chance of accidentally reacting to self.

Matzinger's Danger Theory debates this point of view (for a good introduction, see Matzinger [18]). Technical overviews can be found in Matzinger [19] and Matzinger [20]. She points out that there must be discrimination happening that goes beyond the self-non-self distinction described above. For instance:

- There is no immune reaction to foreign bacteria in the gut or to the food we eat although both are foreign entities.
- Conversely, some auto-reactive processes are useful, for example against self molecules expressed by stressed cells.
- The definition of self is problematic – realistically, self is confined to the subset actually seen by the lymphocytes during maturation.
- The human body changes over its lifetime and thus self changes as well. Therefore, the question arises whether defences against non-self learned early in life might be autoreactive later.
- Other aspects that seem to be at odds with the traditional viewpoint are autoimmune diseases and certain types of tumours that are fought by the immune system (both attacks against self) and successful transplants (no attack against non-self).

Matzinger concludes that the immune system actually discriminates "some self from some non-self". She asserts that the Danger Theory introduces not just new labels, but a way of escaping the semantic difficulties with self and non-self, and thus provides grounding for the immune response. If we accept the Danger Theory as valid we can take care of 'non-self but harmless' and of 'self but harmful' invaders into our system. To see how this is possible, we will have to examine the theory in more detail.

The central idea in the Danger Theory is that the immune system does not respond to non-self but to danger. Thus, just like the self-non-self theories, it fundamentally supports the need for discrimination. However, it differs in the answer to what should be responded to. Instead of responding to foreignness, the immune system reacts to danger.

This theory is borne out of the observation that there is no need to attack everything that is foreign, something that seems to be supported by the counter examples above. In this theory, danger is measured by damage to cells indicated by distress signals that are sent out when cells die an unnatural death (cell stress or lytic cell death, as opposed to programmed cell death, or *apoptosis*).

Figure 1 depicts how we might picture an immune response according to the Danger Theory. A cell that is in distress sends out an alarm signal, whereupon antigens in the neighbourhood are captured by *antigen-presenting cells* such as macrophages, which then travel to the local lymph node and present the antigens to lymphocytes. Essentially, the danger signal establishes a danger zone around itself. Thus B cells producing antibodies that match antigens within the danger zone get stimulated and undergo the clonal expansion process. Those that do not match or are too far away do not get stimulated.

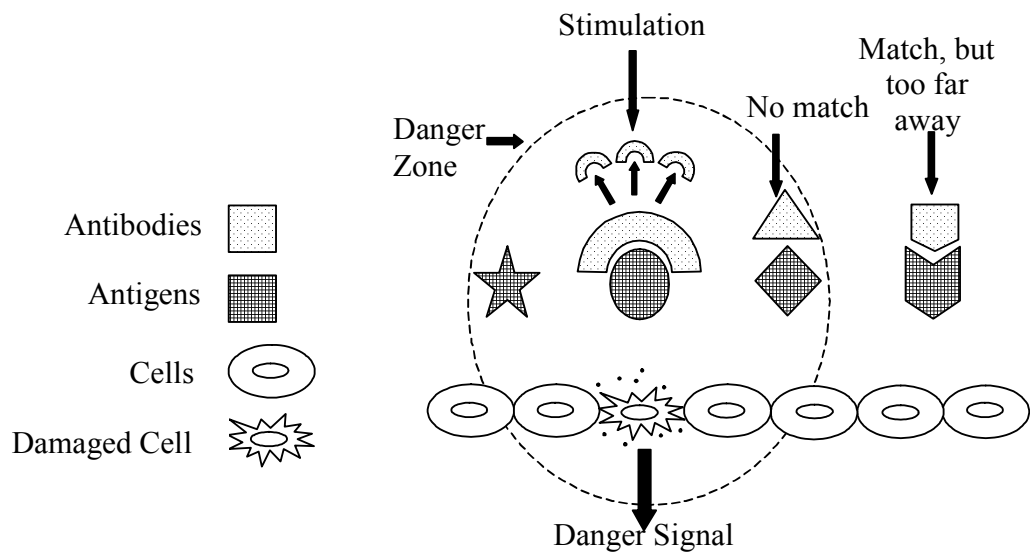

Figure 1: Danger Theory Model.

Matzinger admits that the exact nature of the danger signal is unclear. It may be a 'positive' signal (for example heat shock protein release) or a 'negative' signal (for example lack of synaptic contact with a dendritic antigen-presenting cell). This is where the Danger Theory shares some of the problems associated with traditional self-non-self discrimination (i.e. how to discriminate danger from non-danger). However, in this case, the signal is grounded rather than being some abstract representation of danger.

Another way of looking at the danger model is to see it as an extension of the Two-Signal model by Bretscher and Cohn [4]. In this model, the two signals are antigen recognition (signal one) and co-stimulation (signal two). Co-stimulation is a signal that means "this antigen really is foreign" or, in the Danger Theory, "this antigen really is dangerous". How the signal arises will be explained later. The Danger Theory then operates by applying three laws to lymphocyte behaviour (the *laws of lymphotics* [20]):

- Law 1. Become activated if you receive signals one and two together. Die if you receive signal one in the absence of signal two. Ignore signal two without signal one.

- Law 2. Accept signal two from antigen-presenting cells only (or, for B cells, from T helper cells). B cells can act as antigen-presenting cells only for experienced (memory) T cells. Note that signal one can come from any cells, not just antigen-presenting cells.

- Law 3. After activation (activated cells do not need signal two) revert to resting state after a short time.

For the mature lymphocyte, (whether virgin or experienced) these rules are adhered to. However, there are two exceptions in the lymphocyte lifecycle. Firstly, immature cells are unable to accept signal two from any source. This enables an initial negative selection screening to occur. Secondly, activated (effector) cells respond only to signal one (ignoring signal two), but revert to the resting state shortly afterwards.

An implication of this theory is that autoreactive effects are not necessarily harmful, and are in fact expected during an infection. This is because any lymphocyte reacting to an antigen in the 'danger zone' will be activated. These antigens are not necessarily the culprits for the danger signal. If they are, then the reacting lymphocytes will continue to be restimulated until the antigens (and therefore the danger signal) are removed. After this, they will rest, receiving neither signal one nor signal two.

On the other hand, lymphocytes reacting to innocuous (self) antigens will continue to receive signal one from these antigens, even after the danger (and therefore signal two) has vanished. Therefore these lymphocytes will be deleted, and tolerance will be achieved. However, further autoreactive effects can be expected, partly because 'self' changes over time, and partly because of new lymphocyte generation (particularly B cells, which produce hypermutated clones during activation).

A problem is posed by the antigen-presenting cell itself, whose (innocuous) antigens are by definition always in the danger zone. Lymphocytes reacting to these antigens might destroy the antigen-presenting cell and thus interfere with the immune response. The negative selection of immature lymphocytes protects against this possibility.

Figure 2 shows a more detailed picture of how the Danger Theory can be viewed as an extension of immune signals. These diagrams are adapted from those presented in Matzinger [19] except for the sixth, which incorporates suggestions made in Matzinger [20].

In the original view of the world by Burnet [6], only signal one is considered. This is shown in the first diagram, where the only signal shown is that between infectious agents and lymphocytes (B cells, marked B, and T killer, marked Tk). Signal two (second diagram) was introduced by Bretscher and Cohn [4]. This helper signal comes from a T helper cell (marked Th), on receipt of signal one from the B cell. That is, the B cell presents antigens to the T helper cell and awaits the T cell's confirmation signal. If the T cell recognises the antigen (which, if negative selection has worked, should mean the antigen is non-self) then the immune response can commence. It was Lafferty and Cuningham [17] who proposed that the T helper cells themselves also need to be 'switched on' by signals one and two, both from antigen-presenting cells. This process is depicted in the third diagram.

Note that the T helper cell gets signal one from two sources – the B cell and the antigen-presenting cell. In the former case the antigens are not chosen randomly – the very opposite, since B cells are highly selective for a range of (hopefully non-self) antigens. In the latter case, the antigens are chosen randomly (the antigen-presenting cell simply presents any antigen it picks up) but signal two should only be provided to the T helper cell for non-self antigens. It is not necessarily clear how the antigen-presenting cell 'knows' the antigen is non-self. Janeway [14] introduced the idea of infectious non-self (for example bacteria), which 'primes' antigen presenting cells, i.e. causing signal two to be produced (fourth diagram). This priming signal is labelled as signal 0 in the figures.

Matzinger proposes to allow priming of antigen-presenting cells by a danger signal (fifth diagram). She also proposes to extend the efficacy of T helper cells by routing signal two through antigen presenting cells [20]. We have marked this as 'signal 3' in the sixth diagram (although Matzinger does not use that term, the intention is clear). In Matzinger's words "the antigen seen by the killer need not be the same as the helper; the only requirement is that they must both be presented by the same antigen-presenting cell". This arrangement allows T helper cells to prime many more T killer cells than they would otherwise have been able to.

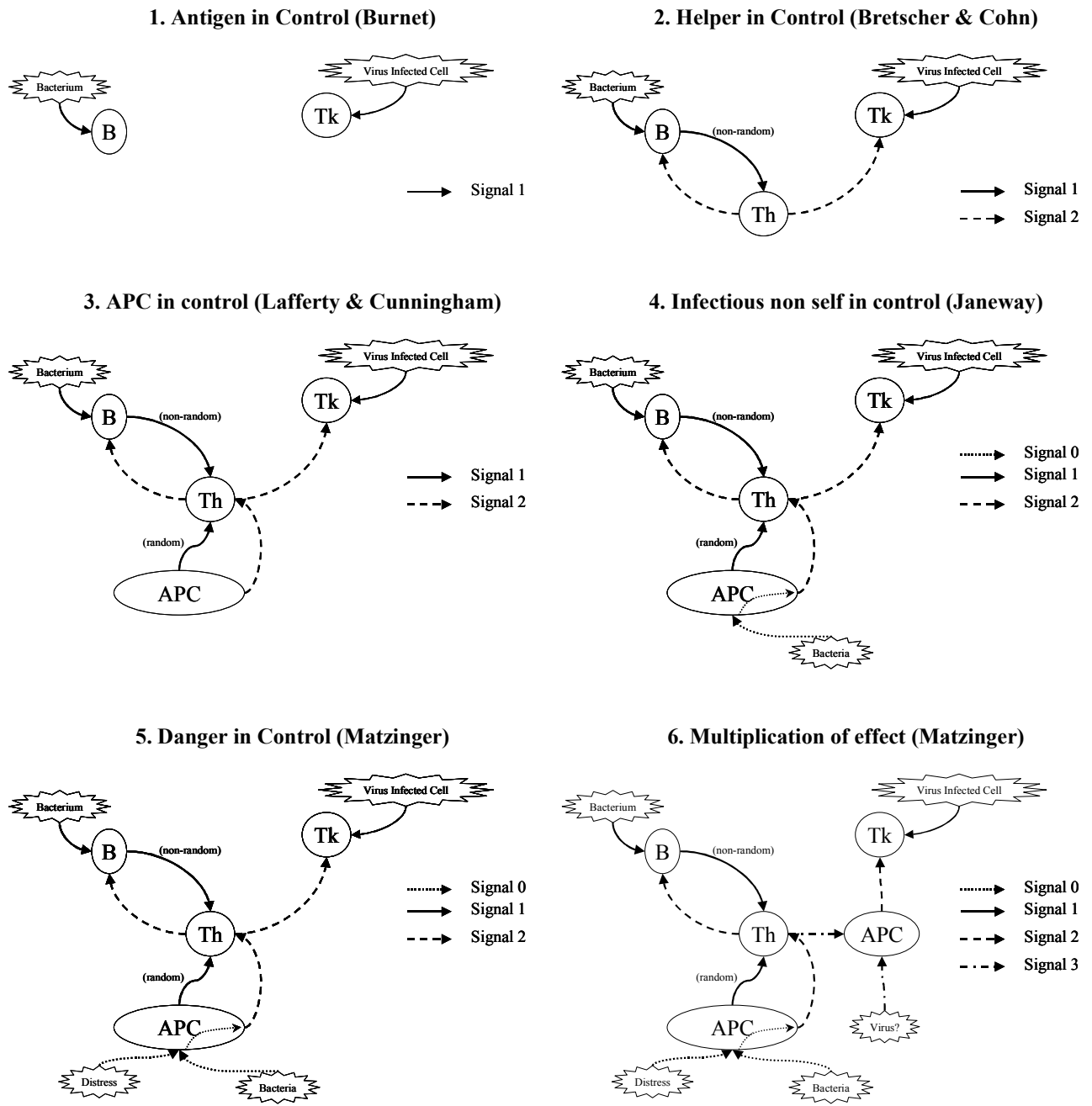

Figure 2: Danger Theory viewed as immune signals.

The Danger Theory is not without its limitations. As mentioned, the exact nature of the danger signal is still unclear. Also, there is sometimes danger that should not be responded to (cuts, transplants). In fact, in the case of transplants it is often necessary to remove the antigen-presenting cells from the transplanted organ. Finally, the fact that autoimmune diseases do still, if rarely, happen, has yet to be fully reconciled with the Danger Theory.

## 3 THE DANGER THEORY AND SOME ANALOGIES TO ARTIFICIAL IMMUNE SYSTEMS

Danger theory clearly has many facets and intricacies, and we have touched on only a few. It might be instructive to list a number of considerations for an Artificial Immune System practitioner regarding the suitability of the danger model for their application. The basic consideration is

whether negative selection is important. If so, then these points may be relevant:

- Negative selection is bound to be imperfect, and therefore autoreactions (false positives) are inevitable.
- The self/non-self boundary is blurred since self and non-self antigens often share common regions.
- Self changes over time. Therefore, one can expect problems with memory cells, which later turn out to be inaccurate or even autoreactive.

If these points are sufficient to make a practitioner consider incorporating the Danger theory into their model, then the following considerations may be instructive:

1. A danger model requires an antigen-presenting cell, which can present an appropriate danger signal.
2. 'Danger' is an emotive term. The signal may have nothing to do with danger (see, for example, our discussion on data mining applications in section 5).
3. The appropriate danger signal can be positive (presence of signal) or negative (absence).
4. The danger zone in biology is spatial. In Artificial Immune System applications, some other measure of proximity (for instance temporal) may be used.
5. If there is an analogue of an immune response, it should not lead to further danger signals. In biology, killer cells cause a normal cell death, not danger.
6. Matzinger proposes priming killer cells via antigen-presenting cells for greater effect. Depending on the immune system used (it only makes sense for spatially distributed models) this proposal may be relevant.
7. There are a variety of considerations that are less directly related to the danger model. For example, migration – how many antibodies receive signal one/two from a given antigen-presenting cell? In addition, the danger theory relies on concentrations, i.e. continuous not binary matching.

There are also a couple of points that might tempt a practitioner to alter the danger model as presented here. For example, the danger model has quite a number of elements. Given that the antigen-presenting cell mediates the danger signal, we might be able to simplify the model – for example, do we still need a T helper cell? In addition, there are some danger signals that might in some sense be 'appropriate' and thus should not trigger an immune response. In such cases, a method for avoiding the danger pathway must be found. A biological example is transplanted organs, in which antigen-presenting cells are removed.

## 4 THE DANGER THEORY AND ANOMALY DETECTION

An intriguing area for the application of Artificial Immune Systems is the detection of anomalies such as computer viruses, fraudulent transactions or hardware faults. The underlying metaphor seems to fit particularly nicely here, as there is a system (self) that has to be protected against intruders (non-self). Thus if natural immune systems have enabled biological species to survive, can we not create Artificial Immune Systems to do the same to our computers, machines etc? Presumably those systems would then have the same beneficial properties as natural immune systems like error tolerance, distribution, adaptation and self-monitoring. A recent overview of biologically inspired approaches to this area can be found in Williamson [22].

In this section we will present indicative examples of such artificial systems, explain their current shortcomings and show how the Danger Theory might help overcome some of these.

One of the first such approaches is presented by Forrest et al [11] and extended by Hofmeyr and Forrest [13]. This work is concerned with building an Artificial Immune System that is able to detect non-self in the area of network security where non-self is defined as an undesired connection. All connections are modelled as binary strings and there is a set of known good and bad connections, which is used to train and evaluate the algorithm. To build the Artificial Immune System, random binary strings are created called detectors.

These detectors then undergo a maturation phase where they are presented with good, i.e. self, connections. If they match any of these they are eliminated otherwise they become mature, but not activated. If during their further lifetime these mature detectors match anything else, exceeding a certain threshold value, they become activated. This is then reported to a human operator who decides whether there is a true anomaly. If so the detectors are promoted to memory detectors with an indefinite life span and minimum activation threshold. Thus, this is similar to the secondary response in the natural immune system, for instance after immunisation.

An approach such as the above is known in Artificial Immune Systems as negative selection as only those detectors (antibodies) that do not match live on. It is thought that T cells mature in similar fashion in the thymus such that only those survive and mature that do not match any self cells after a certain amount of time.

An alternative approach to negative selection is that of positive selection as used for instance by Forrest et al [9] and by Somayaji and Forrest [22]. These systems are a reversal of the negative selection algorithm described above with the difference that detectors for self are evolved. From a performance point of view there are advantages and disadvantages for both methods. A suspect non-self string would have to be compared with all self-detectors to establish that it is non-self, whilst with negative selection the first matching detector would stop the comparison. On the other hand, for a self-string this is reversed giving positive selection the upper hand. Thus, performance depends on the self to non-self ratio, which should generally favour positive selection.

However, there is another difference between the two approaches: the nature of false alarms. With negative selection inadequate detectors will result in false negatives (missed intrusions) whilst with positive selection there will be false positives (false alarms). The preference between the two in this case is likely to be problem specific.

Both approaches have been extended further [10] including better co-stimulation methods and activation thresholds to reduce the number of false alarms, multiple antibody sub-populations for improved diversity and coverage and improved partial matching rules. Recently, similar approaches have also been used to detect hardware faults (Bradley and Tyrrell [1]), network intrusion (Kim and Bentley [16]) and fault tolerance (Burgess [5]).

What are the remaining challenges for a successful use of Artificial Immune Systems for anomaly detection? Firstly, self and non-self will usually evolve and change during the lifetime of the system. Hence, to be effective, any system used must be robust and flexible enough to cope with changing circumstances. Based on the performance of their natural counterparts, Artificial Immune Systems should be well suited to provide these qualities. Secondly, appropriate representations of self and good matching rules have to be developed. Most research so far has been concentrated in these two areas and good advances have been made so far [8].

However, as pointed out by Kim and Bentley [15], scaling is a problem with negative selection. As the systems to be protected grow larger and larger so does self and non-self and it becomes more and more problematic to find a set of detectors that provides adequate coverage whilst being computationally efficient. It is inefficient, if not impossible, to map the entire non-self universe, particularly as it will be changing over time. The same applies to positive selection and trying to map all of self.

Moreover, the approaches so far have another disadvantage: A response requires infection beyond a certain threshold and human intervention confirming this. Although one might argue that the operator sees fewer alarms than in an unaided system, this clearly is not yet the ideal situation of an autonomous system preventing all damage. Apart from the resource implication of a human component, an unduly long delay might be caused by this necessity prolonging the time the system is exposed. This situation might be further aggravated by the fact that the labels self and non-self are often ambiguous and expert knowledge might be required to apply them correctly.

How can these problems be overcome? We believe that applying ideas from the Danger Theory can help building better Artificial Immune Systems by providing a different way of grounding and removing the necessity to map self or non-self. To achieve this self-non-self discrimination will still be useful but it is no longer essential. This is because non-self no longer causes an immune response. Instead, it will be danger signals that trigger a reaction.

What could such danger signals be? They should show up after limited infection to minimise damage and hence have to be quickly and automatically measurable. Suitable signals could include:

- Too low or too high memory usage.
- Inappropriate disk activity.
- Unexpected frequency of file changes as measured for example by checksums or file size.
- SIGABRT signal from abnormally terminated UNIX processes.
- Presence of non-self.

Of course, it would also be possible to use 'positive' signals, as discussed in the previous section, such as the absence of some normal 'health' signals.

Once the danger signal has been transmitted, the immune system can then react to those antigens, for example, executables or connections, which are 'near' the emitter of the danger signal. Note that 'near' does not necessarily mean geographical or physical closeness, something that might make sense for connections and their IP addresses but probably not for computer executables in general. In essence, the physical 'near' that the Danger Theory requires for the immune system is a proxy measure for causality. Hence, we can substitute it with more appropriate causality measures such as similar execution start times, concurrent runtimes or access of the same resources.

Consequently, those antibodies or detectors that match (first signal) those antigens within a radius, defined by a measure such as the above (second signal), will proliferate. Having thereby identified the dangerous components, further confirmation could then be sought by sending it to a special part of the system simulating another attack. This would have the further advantage of not having to send all detectors to confirm danger. In conclusion, using these ideas from the Danger Theory has provided a better grounding of danger labels in comparison to self / non-self, whilst at the same time relying less on human competence.

## 5 THE DANGER THEORY AND OTHER ARTIFICIAL IMMUNE SYSTEM APPLICATIONS

It is not immediately obvious how the Danger Theory could be of use to data mining problems such as the movie prediction problem described in Cayzer and Aickelin [7], because the notions of self and non-self are not used. In essence, in data mining all of the system is self. More precisely, it is not an issue what is self or non-self as the designer of the database has complete control over this aspect.

However, if the labels self and non-self were to be replaced by interesting and non-interesting data for example, a distinction would prove beneficial. In this case, the immune system is being applied as a classifier. If one can then further assume that interesting data is located 'close' or 'near' to other interesting data, ideas from the Danger Theory can come into play again. To do so, it is necessary to define 'close' / 'near'. We could use:

- Physical closeness, for instance distance in the database as measured by an appropriate metric.
- Correlation of data, as measured by statistical tools.
- Similar entry times into the database.
- File size.

A danger signal could thus be interpreted as a valuable piece of information that has been uncovered. Hence, those antibodies are stimulated that match data that is 'close' this valuable piece of information.

Taking this idea further, we might define the danger signal as an indication of user interest. Given this definition, we can speculate about various scenarios in which the danger signal could be of use. One such scenario is outlined below for illustrative purposes.

Imagine a user browsing a set of documents. Each document has a set of features (for instance keywords, title, author, date etc). Imagine further that there is an immune system implemented as a 'watcher', whose antibodies match document features. 'Interesting' documents are those, whose features are matched by the immune system.

When a user either explicitly or implicitly indicates interest in the current document, a "danger" signal is raised. This causes signal two to be passed, along with signal one, to antibodies matching any antigen, i.e. document feature, in the danger zone, i.e. this document.

Stimulated antibodies become effectors, and thus the immune system learns to become a good filter when searching for other interesting documents. Interesting documents could be brought to the user's attention (the exact mechanism is not relevant here). The important thing is that the user's idea of an 'interesting' document may change over time and so it is important that the immune system adapts in a timely way to such a changing definition of (non-) self.

Meanwhile, every document browsed by the user (whether interesting or not) will be presented to the antibodies as 'signal one'. Uninteresting document features will therefore give rise to signal one without signal two, which will tolerate the autoreactive antibodies. The net effect is to produce a set of antibodies that match only interesting document features.

As mentioned, this example is purely illustrative but it does show that ideas from the Danger theory may have implications for Artificial Immune System applications in domains where the relevance of 'danger' is far from obvious.

# 6 CONCLUSIONS

To conclude, the Danger Theory is not about the way Artificial Immune Systems represent data. Instead, it provides ideas about which data the Artificial Immune Systems should represent and deal with. They should focus on dangerous, i.e. interesting data.

It could be argued that the shift from non-self to danger is merely a symbolic label change that achieves nothing. We do not believe this to be the case, since danger is a grounded signal, and non-self is (typically) a set of feature vectors with no further information about their meaning. The danger signal helps us to identify which subset of feature vectors is of interest. A suitably defined danger signal thus overcomes many of the limitations of self-non-self selection. It restricts the domain of non-self to a manageable size, removes the need to screen against all self, and deals adaptively with scenarios where self (or non-self) changes over time.

The challenge is clearly to define a suitable danger signal, a choice that might prove as critical as the choice of fitness function for an evolutionary algorithm. In addition, the physical distance in the biological system should be translated into a suitable proxy measure for similarity or causality in an Artificial Immune System. We have made some suggestions in this paper about how to tackle these challenges in a variety of domains, but the process is not likely to be trivial. Nevertheless, if these challenges are met, then future Artificial Immune System applications might derive considerable benefit, and new insights, from the Danger Theory.

**Acknowledgements**

We would like to thank the two anonymous reviewers, whose comments greatly improved this paper.

**References**

[1] Bradley D, Tyrell A, The Architecture for a Hardware Immune System, *Proceedings of the 2002 Congress on Evolutionary Computation*, 2002.

[2] Forrest H. Bennett III, John R. Koza, Jessen Yu, William Mydlowec, Automatic Synthesis, Placement, and Routing of an Amplifier Circuit by Means of Genetic Programming. *Evolvable Systems: From Biology to Hardware, Third International Conference, ICES 2000:* 1-10, 2000

[3] D. W. Bradley, Andrew M. Tyrrell, Immunotronics: Hardware Fault Tolerance Inspired by the Immune System. *Evolvable Systems: From Biology to Hardware, Third International Conference, ICES 2000:* 11-20, 2000

[4] Bretscher P, Cohn M, A theory of self-nonself discrimination, *Science* 169, 1042-1049, 1970

[5] Burgess M: Computer Immunology, *Proceedings of LISA XII*, 283-297, 1998.

[6] Burnet F, *The Clonal Selection Theory of Acquired Immunity*, Vanderbilt University Press, Nashville, TN, 1959.

[7] Cayzer S, Aickelin U, A Recommender System based on the Immune Network, *Proceedings of the 2002 Congress on Evolutionary Computation*, 2002.

[8] Dasgupta D, Majumdar N, Nino F, Artificial Immune Systems: A Bibliography, *Computer Science Division, University of Memphis, Technical Report* No. CS-02-001, 2001.

[9] Forrest S, Hofmeyr S, Somayaji A, Longstaff T, A sense of self for Unix processes, *Proceedings of the 1996 IEEE Symposium on Research in Security and Privacy*, 120-128, 1996.

[10] Forrest S, http://www.cs.unm.edu/~immsec/, 2002.

[11] Forrest S, Perelson A, Allen L, Cherukuri R, Self-non-self discrimination in a computer, *Proceedings of the 1994 IEEE Symposium on Research in Security and Privacy*, 202-212, 1994.

[12] Goldsby R, Kindt T, Osborne B, *Kuby Immunology*, Fourth Edition, W H Freeman, 2000.

[13] Hofmeyr S, Forrest S, Architecture for an Artificial Immune System, *Evolutionary Computation* 8(4), 443-473, 2000.

[14] Janeway C, The immune System evolved to discriminated infectious nonself from noninfectious self, *Immunology Today* 13, 11-16, 1992.

[15] Kim J, Bentley P, An evaluation of negative selection in an artificial immune for network intrusion detection, *Proceedings of the 2001 Genetic and Evolutionary Computation Conference*, 1330-1337, 2001.

[16] Kim J, Bentley P, Towards an Artificial Immune System for Network Intrusion Detection: An Investigation of Dynamic Clonal Selection. *Proceedings of the 2002 Congress on Evolutionary Computation*, 2002.

[17] Lafferty K, Cunningham A, A new analysis of allogeneic interactions. *Australian Journal of Experimental Biology and Medical Sciences*. 53:27-42, 1975

[18] Matzinger P, http://cmmg.biosci.wayne.edu/asg/polly.html

[19] Matzinger P, The Danger Model in Its Historical Context, *Scandinavian Journal of Immunology*, 54: 4-9, 2001.

[20] Matzinger P, Tolerance, Danger and the Extended Family, *Annual Review of Immunology*, 12:991-1045, 1994.

[21] Langman R (editor), Self non-self discrimination revisited, *Seminars in Immunology* 12, Issue 3, 2000.

[22] Somayaji A, Forrest S, Automated response using system-call delays, *Proceedings of the ninth USENIX Security Symposium*, 185-197, 2000.

[23] Williamson M, Biologically inspired approaches to computer security, HP Labs Technical Reports HPL-2002-131, 2000 (available from http://www.hpl.hp.com/techreports/2002/HPL-2002-131.html).